\documentclass{article}

\usepackage[preprint]{neurips_2023}

\usepackage{graphicx}
\usepackage{float}
\usepackage{makecell}
\usepackage{amsmath}
\usepackage{bm}
\usepackage{gensymb}
\usepackage{subcaption}
\usepackage[utf8]{inputenc} 
\usepackage[T1]{fontenc}    
\usepackage{hyperref}       
\usepackage{url}            
\usepackage{booktabs}       
\usepackage{amsfonts}       
\usepackage{nicefrac}       
\usepackage{microtype}      
\usepackage{xcolor}         

\title{HEAL-ViT: Vision Transformers on a spherical mesh for medium-range weather forecasting }

\author{%
  Vivek Ramavajjala\thanks{https://excarta.io} \\
  Excarta, Inc.\\
  San Carlos, CA 15213 \\
  \texttt{vivek@excarta.io} \\
}

\begin{document}

\maketitle

\begin{abstract}
  In recent years, a variety of ML architectures and techniques have seen success in producing skillful medium range weather forecasts. In particular, Vision Transformer (ViT)-based models (e.g. Pangu-Weather, FuXi) have shown strong performance, working nearly "out-of-the-box" by treating weather data as a multi-channel image on a rectilinear grid.
  While a rectilinear grid is appropriate for 2D images, weather data is inherently spherical and thus heavily distorted at the poles on a rectilinear grid, leading to disproportionate compute being used to model data near the poles.
  Graph-based methods (e.g. GraphCast) do not suffer from this problem, as they map the longitude-latitude grid to a spherical mesh, but are generally more memory intensive and tend to need more compute resources for training and inference. While spatially homogeneous, the spherical mesh does not lend itself readily to be modeled by ViT-based models that implicitly rely on the rectilinear grid structure. We present HEAL-ViT, a novel architecture that uses ViT models on a spherical mesh, thus benefiting from both the spatial homogeneity enjoyed by graph-based models and efficient attention-based mechanisms exploited by transformers. HEAL-ViT produces weather forecasts that outperform the ECMWF IFS on key metrics, and demonstrate better bias accumulation and blurring than other ML weather prediction models. Further, the lowered compute footprint of HEAL-ViT makes it attractive for operational use as well, where other models in addition to a 6-hourly prediction model may be needed to produce the full set of operational forecasts required.
\end{abstract}

\section{Introduction}
In recent years, a variety of ML Weather Prediction models (MLWPs) have demonstrated strong
performance in medium-range weather prediction, defined as task of producing 10 day
forecasts from a given initialization condition. MLWPs have typically been trained on ECMWF's
ERA5\cite{hersbach2020era5} archive and have shown improvements in key metrics over the ECMWF IFS model, which is
generally considered the state-of-the-art in Numerical Weather Prediction (NWP)\cite{haiden2018evaluation}. A variety of
model architectures, with prominent ones being FourCastNet\cite{pathak2022fourcastnet}, Pangu-Weather\cite{bi2023accurate}, GraphCast\cite{lam2022graphcast}, and FuXi\cite{chen2023fuxi},
have been successful at producing high-quality 10 day forecasts at the native 0.25\textdegree
resolution provided by the ERA5\cite{hersbach2020era5} dataset. While all MLWPs show reductions in RMSE over the IFS baseline, they tend to accumulate biases more strongly and produce blurrier forecasts (as measured by zonal energy spectra) than IFS. Depending on the choice of metric and variable, different MLWPs have better or worse performance, e.g., while GraphCast performs better with RMSE as a metric for most variables, vision transformer (ViT)\cite{50650} based models like Pangu-Weather tend to show reduced blurring and do not accumulate biases as strongly.\par

All architectures use an "encoder - processor - decoder" scheme, where (i) the encoder maps the
original 0.25\textdegree longitude-latitude grid to a smaller mesh, (ii) the processor models
interactions between nodes on the smaller mesh, and (iii) the decoder maps the processed mesh
back to the longitude-latitude grid. FourCastNet, Pangu-Weather, and FuXi use convolutional
layers in the encoder and decoder, mapping the longitude-latitude grid to a rectilinear mesh. GraphCast
uses graph networks to map the grid to a icosahedral mesh which distributes
nodes evenly on a sphere. For the processor, FourCastNet uses Fourier neural operators, while
Pangu-Weather and FuXi employ ViT architectures, all operating on
a rectilinear mesh. GraphCast employs graph networks for message passing between nodes on the
spherical icosahedral mesh. While ViT-based approaches are exploit attention mechanisms
to effectively learn long-distance relationships in the mesh, the rectilinear mesh used over-represents data near the
poles, thus needing disproportionate compute resources to model the spatially inhomogenous
representation. While graph-based approaches operate on spatially homogenous representations,
they tend to be more memory intensive owing to the message-passing mechanisms required.

\subsection{Our contribution}
We present HEAL-ViT, a novel architecture that allows standard vision transformers
to be used on a spherical mesh, thus benefiting from the uniform spatial representation offered
by spherical meshes and from attention mechanisms that can model long-distance relationships in
a memory-efficient manner. HEAL-ViT has a lower memory and compute footprint as the spatially homogenous spherical mesh is more efficient at representing the underlying data, making it attractive for operational use where a combination of models may be needed to produce operational weather forecasts. HEAL-ViT is comparable in performance with other MLWPs in
terms of error reduction, but shows improved performance in both bias accumulation and blurring,
as measured by zonal energy spectra.

\subsection{Outline}
After defining the problem, we provide a brief overview of SWIN transformer models, a general-purpose variant of vision transformers that operate on rectilinear grids. We then describe the HEALPix mesh and how it can be adapted for use with SWIN transformers. Subsequent sections describe the HEAL-ViT model architecture, network parameterization, training details, and finally results.

\section{Problem definition and notation}
\label{section:problem}
Using $Z^t$ to denote the true weather state at time $t$, the problem of weather forecasting can be defined as finding the transition function $\Phi$ that predicts the future state $Z^{t+\Delta t} = \Phi(Z^{t})$. As $Z^t$ encapsulates the complete weather state at time $t$, it contains sufficient information for the task of predicting $Z^{t+\Delta t}$. In practice, only a partial observation $X^{t}$ of the true weather $Z^{t}$ is available, and our objective is to learn the best possible approximation $\phi$ of the true transition function $\Phi$. As $X^t$ is a partial view of the true state $Z^t$, we provide additional historical context in the form $X^{t-\Delta t}$ to learn an accurate $\phi$:
\begin{equation}
  X^{t+\Delta t} = \phi(X^{t}, X^{t-\Delta t})
\end{equation}
Having learned an accurate model $\phi$, given a starting state $X^{t_0}$, we can produce a forecast for a desired length using $\phi$ auto-regressively:
\begin{equation}
  X^{t_0+k\Delta t} = \phi(X^{t_0 + (k-1)\Delta t}, X^{t_0 + (k-2)\Delta t})
\end{equation}
Following other MLWPs, we fix $\Delta t=6$ hours, limit our forecasts to a maximum of 10 days (40 steps), and simplify the notation of $X^{t+k\Delta t}$ to $X^{t+k}$. In our work, the observed weather state $X^t$ consists of the surface and atmospheric variables listed in Table \ref{tab:weatherfeats}. Besides the observed weather states $X^t$ and $X^{t-1}$, we also use as input features the land-sea mask (a binary variable) and surface geopotential (which captures orography). In total, this means $\phi$ consumes 110 features as input: 54 features each for $X^t$ and $X^{t-1}$ and 2 static features.
\begin{table}[ht]
  \centering
  \begin{tabular}{|p{0.4\linewidth} |p{0.3\linewidth} | p{0.2\linewidth} |}
  \hline
  \textbf{Surface variables (9)} & \textbf{Atmospheric variables (5)} & \textbf{Pressure levels} \\
  \hline
  \makecell[tl]{2-meter temperature (\texttt{t2m}) \\
  2-meter dewpoint (\texttt{d2m}) \\
  10-meter u wind compomnent (\texttt{u10}) \\
  10-meter v wind compomnent (\texttt{v10}) \\
  Mean sea-level pressure (\texttt{msl})\\
  Surface pressure (\texttt{sp})\\
  100-meter u wind compomnent (\texttt{u100}) \\
  100-meter v wind compomnent (\texttt{v100}) \\
  Total column water vapour (\texttt{tcwv})}
  &
  \makecell[tl]{Temperature (\texttt{t})\\
  U component of wind (\texttt{u})\\
  V component of wind (\texttt{v})\\
  Geopotential (\texttt{z})\\
  Relative humidity (\texttt{r})}
  &
  \makecell[tl]{1000, 925, 850,\\700, 500, 300,\\250, 200, 50}\\
  \hline
  \end{tabular}
  \caption{Weather variables modeled by HEAL-ViT.}
  \label{tab:weatherfeats}
\end{table}

\section{SWIN transformers on the HEALPix mesh}
\subsection{SWIN transformers}
\begin{figure}
  \centering
  \includegraphics[width=0.8\linewidth]{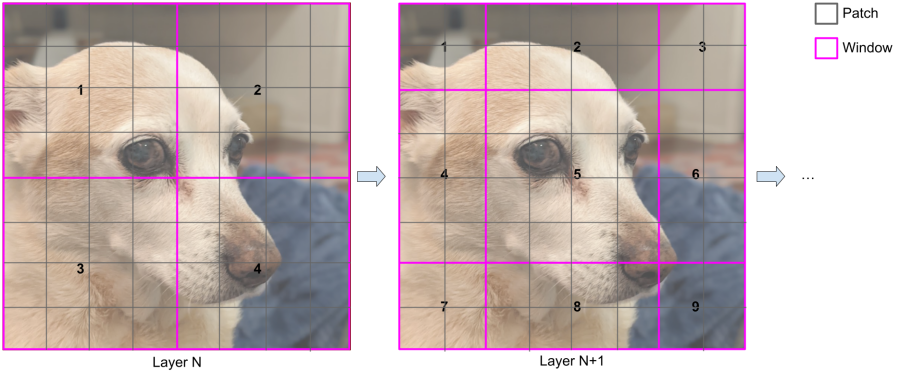}
  \caption{SWIN transformers perform local attention within non-overlapping windows, and in
  alternate layers shift the windows by half the window width to allow for cross-window
  connections to be learned.}
 \label{fig:swinfig}
\end{figure}

SWIN (\textbf{S}hifted \textbf{Win}dows) transformers\cite{Liu_2021_ICCV} are a general-purpose backbone for computer vision tasks, and have been used extensively for medium range weather forecasting\cite{chen2023swinrdm, hu2023swinvrnn,bi2023accurate,chen2023fuxi}. In SWIN transformers, the image is first "patched" into $M \times N$ patches, effectively creating a rectilinear mesh where each mesh node represents a specific part of the input image. Patches are grouped into windows, and self-attention is performed within each window to model dependencies between patches. Cross-window connections are learned by shifting the windows in alternate layers, as illustrated in figure \ref{fig:swinfig}.\par
Performing self-attention only within windows limits the computational cost of the attention mechanism (which grows quadratically with the size of the window), while shifting the windows allows for long-distance relationships to be learned. Shifting the windows is trivial with a rectilinear mesh where all rows and columns have equal number of nodes, but pose a challenge on a spherical mesh where fewer nodes represent data near the poles than the equator. We exploit the favourable properties of the HEALPix pixelation scheme to efficiently shift windows on the spherical mesh and allow SWIN transformers to be used without any modifications.
\subsection{The HEALPix mesh}
HEALPix\cite{gorski1999healpix} (\textbf{H}ierarchical \textbf{E}qual \textbf{A}rea iso\textbf{L}atitude \textbf{Pix}elation of a sphere) is a pixelation scheme that
divides a spherical surface into pixels representing curvilinear quadrilateral
sections of equal area. Originally developed for mapping and analyzing the cosmic microwave background, HEALPix meshes have also been used with spherical CNNs\cite{perraudin2019deepsphere}, \cite{krachmalnicoff2019convolutional}.\par
\begin{figure*}[t!]
    \centering
    \begin{subfigure}[t]{0.4\textwidth}
        \centering
        \includegraphics[width=\textwidth]{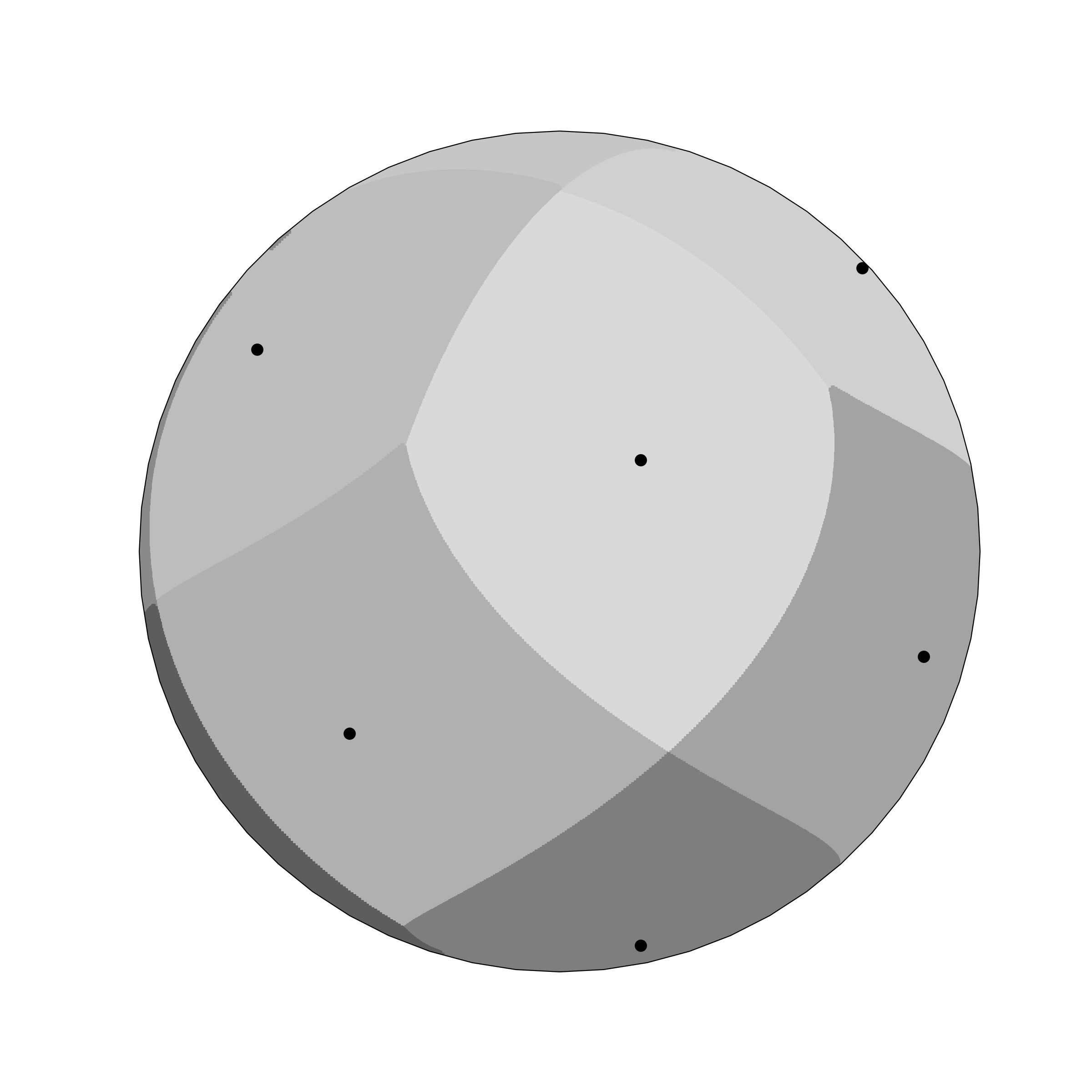}
        \caption{Base HEALPix mesh, $n=0$, 12 pixels. Each face is represented by 1 pixel.}
    \end{subfigure}%
    ~ 
    \begin{subfigure}[t]{0.4\textwidth}
        \centering
        \includegraphics[width=\textwidth]{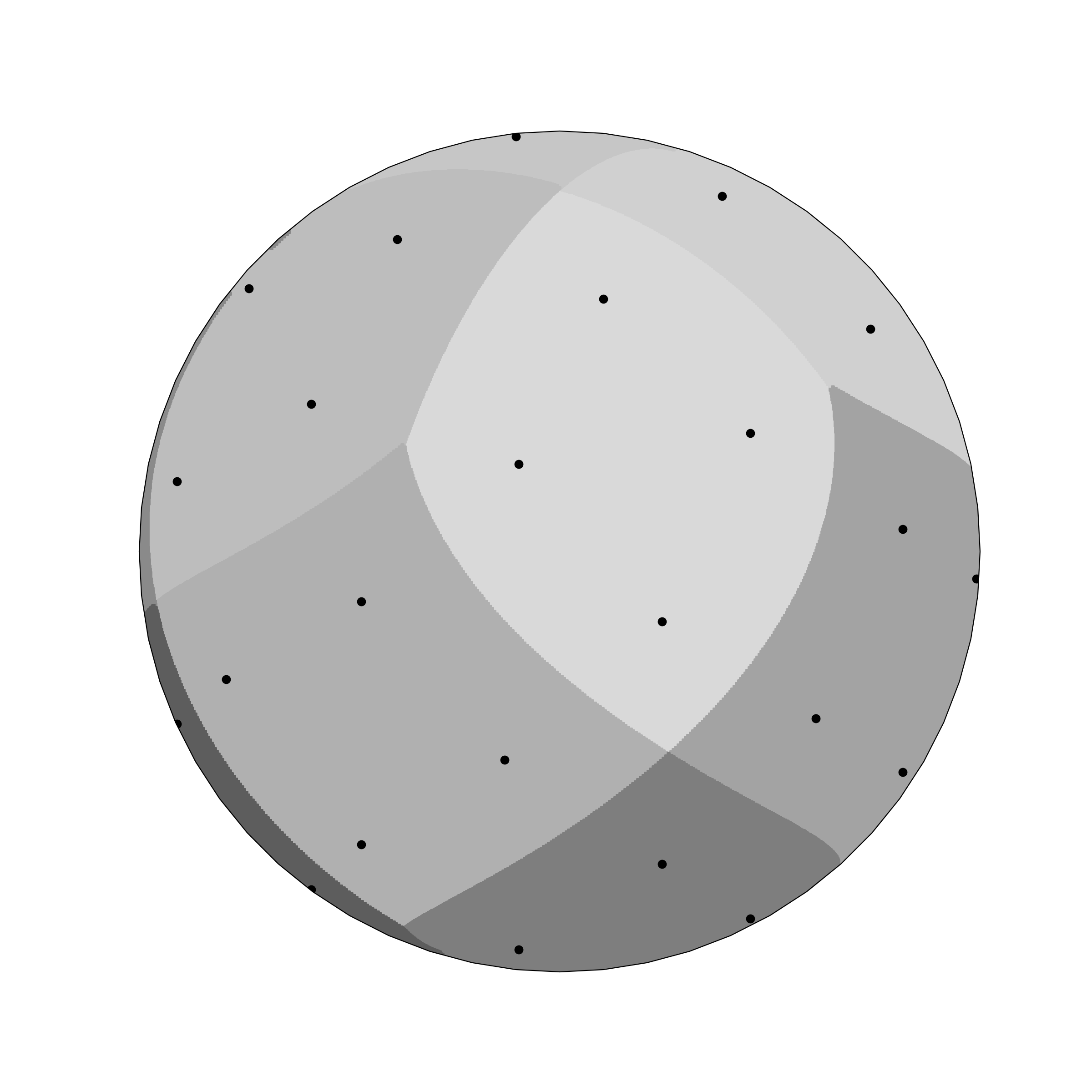}
        \caption{HEALPix mesh, $n=1$, 48 pixels. Each face is represented by 2x2 pixels.}
    \end{subfigure}
    \begin{subfigure}[t]{0.4\textwidth}
        \centering
        \includegraphics[width=\textwidth]{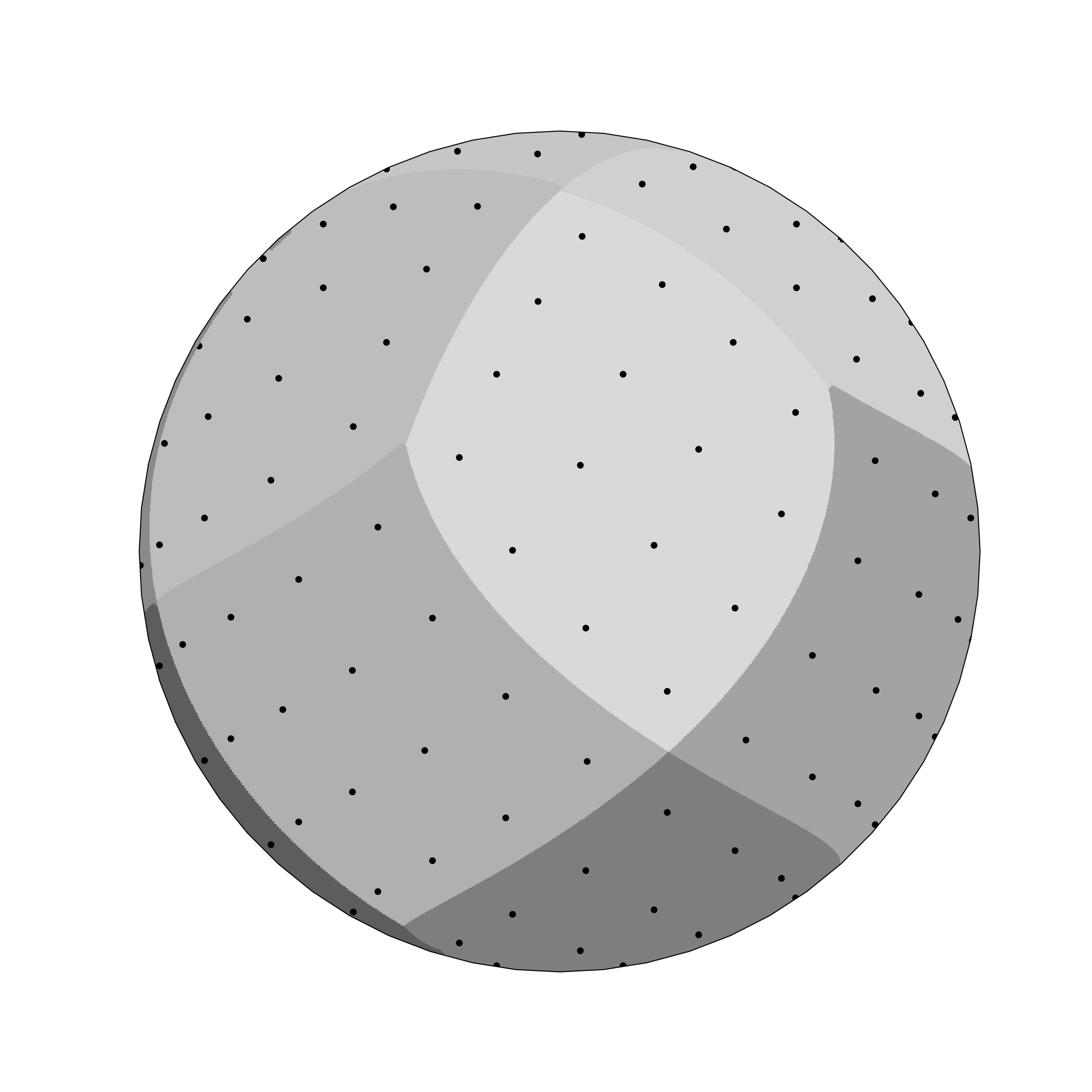}
        \caption{HEALPix mesh, $n=2$, 192 pixels. Each face is represented by 4x4 pixels.}
    \end{subfigure}%
    ~ 
    \begin{subfigure}[t]{0.4\textwidth}
        \centering
        \includegraphics[width=\textwidth]{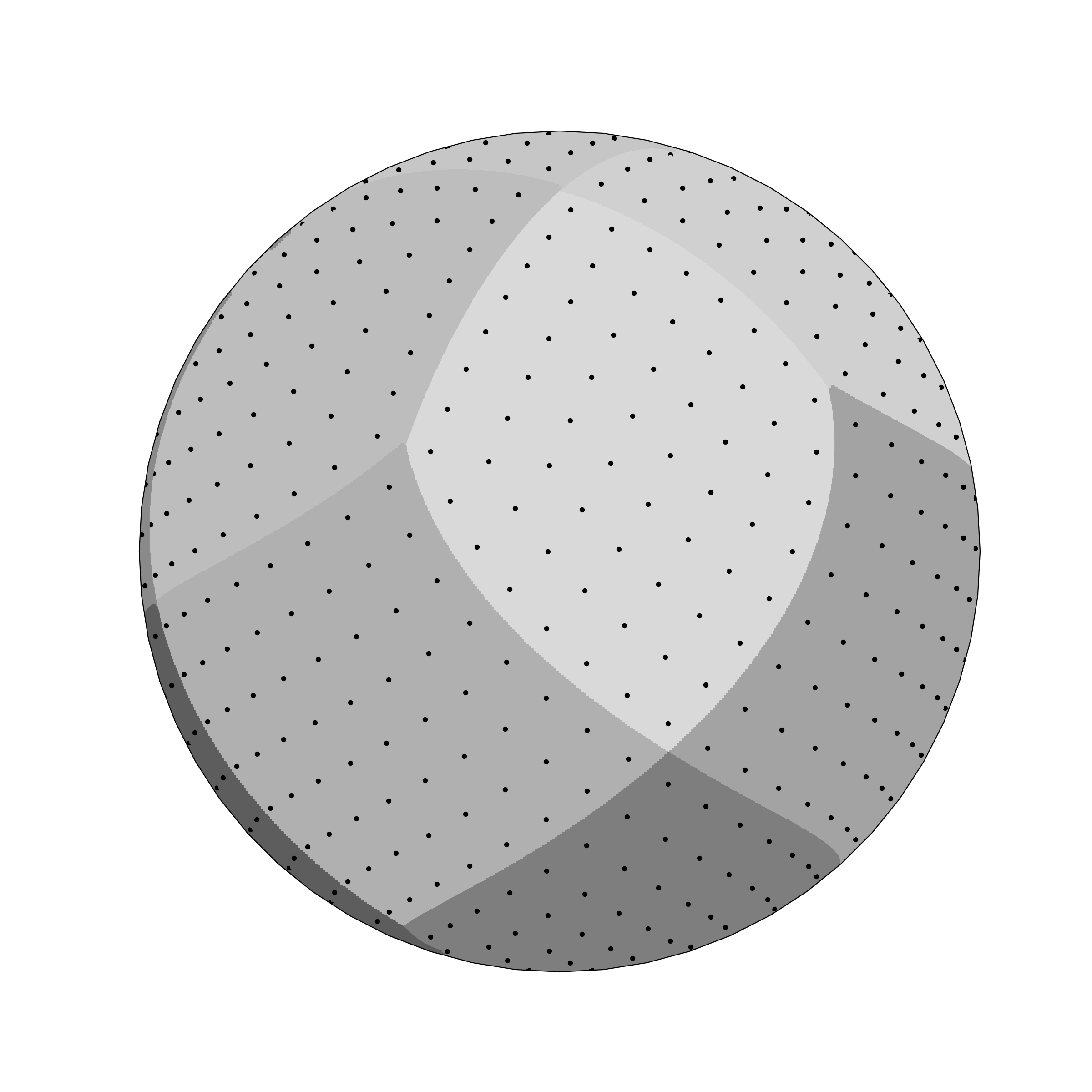}
        \caption{HEALPix mesh, $n=3$, 768 pixels. Each face is represented by 8x8 pixels.}
    \end{subfigure}
    \caption{HEALPix meshes at increasingly finer resolutions.}
    \label{fig:healpix}
\end{figure*}

At the coarsest level, the HEALPix mesh has 12 pixels representing 12 quadrilateral "faces". The mesh is refined by dividing each pixel into 4 sub-pixels that represent the quadrants of the original quadrilateral. Repeating this process $n$ times divides a face into $(2^n)^2$ pixels, arranged as a square of size $2^n \times 2^n$ pixels. The total number of pixels at any level of refinement can be calculated as $12 * ( (2^n)^2)$.
Figure \ref{fig:healpix} illustrates the refinement process for 0, 1, 2, and 3 levels of
refinement, and table \ref{tab:numnodes} shows how the number of nodes increases as a function of refinement level.
\begin{table}[ht]
  \centering
  \begin{tabular}{|l |c |c|c|c|c|c|c|c|}
  \hline
  \textbf{Refinement level} & 0 & 1 & 2 & 3 & 4 & 5 & 6 & 7\\
  \hline
  \textbf{Total number of nodes} & 12 & 48 & 192 & 768 & 3,072 & 12,288 & 49,152 & 196,608 \\
  \hline
  \end{tabular}
  \caption{Number of nodes in a HEALPix mesh as a function of refinement level.}
  \label{tab:numnodes}
\end{table}

\begin{figure}
    \centering
    \includegraphics[width=0.9\linewidth]{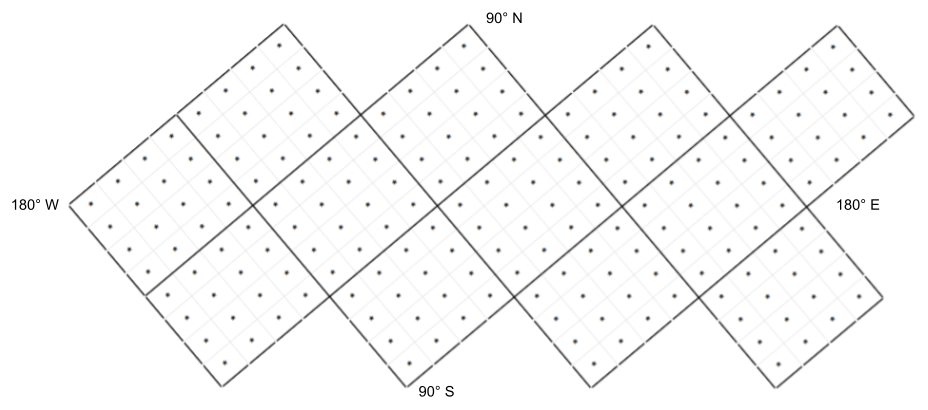}
    \caption{Flattened view of the 12 faces of the HEALPix mesh, each sub-divided into 16 pixels. The HEALPix mesh provides a regular ordering of the mesh nodes, with each mesh node having neighbors along cardinal directions. The whole mesh can thus be oriented North-South and East-West. 4 of the faces are "central", along the equator, while 4 cover the northern hemisphere, and 4 cover the southern hemisphere.}
    \label{fig:hpflatten}
\end{figure}

The HEALPix mesh provides a regular ordering of the mesh nodes, with each mesh node having well-defined neighbors along the East, West, North, South, North-East, North-West, South-East, and South-West directions. As a result, the whole mesh can be oriented North-South and East-West and be "flattened" as seen in \ref{fig:hpflatten}.
\subsubsection{Creating windows on the HEALPix mesh}
\begin{figure}
    \centering
    \includegraphics[width=0.9\linewidth]{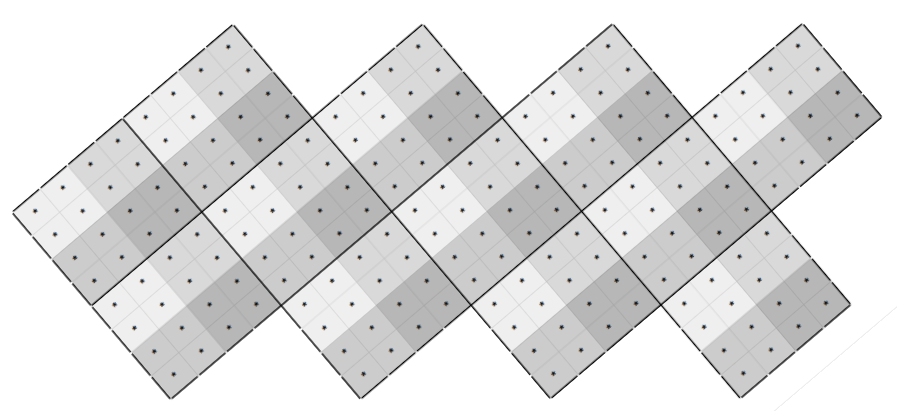}
    \caption{Windowed HEALPix mesh, with each HEALPix face containing 4 windows, and each window containing 4 pixels. This corresponds to a refinement level of $n=2$ with 192 fine pixels, and a window size parameter of $w=1$. Each window contains $(2^w)^2=4$ pixels.}
    \label{fig:hpowindow}
\end{figure}

Each HEALPix mesh node corresponds to a specific part of the sphere, and thus is equivalent to the image "patches" used in SWIN transformers. We can use the hierarchical subdivision of pixels to create the windows needed by SWIN transformers for local self-attention. Given a fine refinement level $n$ and a coarser refinement level $n-w$, we see that each coarse pixel "contains" $(2^w)^2$ fine pixels, and each fine pixel belongs to exactly one coarse pixel. Thus, pixels from a coarse refinement level can be used as non-overlapping windows within which self-attention can be peformed on fine pixels. Figure \ref{fig:hpowindow} shows the layout of windows on a flattened HEALPix mesh. Given a fine mesh at refinement level $n$, the window size parameter $w$ can be tuned to have each fine pixel attend on a larger or smaller context on the sphere.
\subsubsection{Shifting windows on the HEALPix mesh}
\begin{figure}
    \centering
    \includegraphics[width=0.9\linewidth]{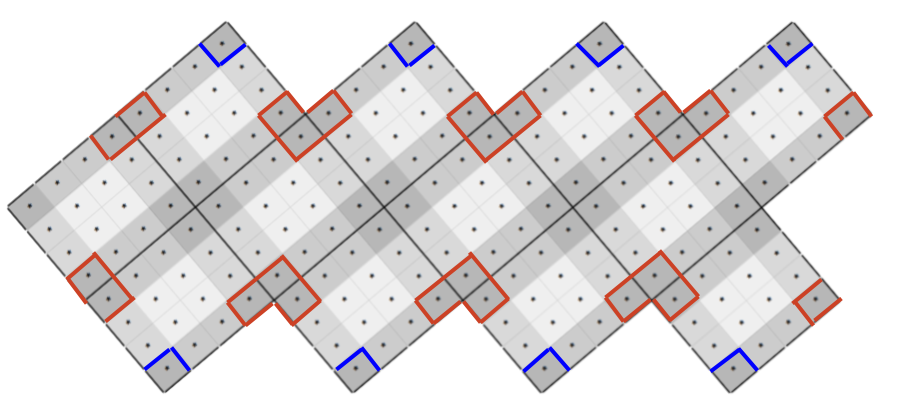}
    \caption{Shifted windows on the HEALPix mesh, where each original window from figure \ref{fig:hpowindow} is effectively shifted by one quadrant south and west.}
    \label{fig:hpnwindow}
\end{figure}

Hierarchical subdivision and equal-area pixels are properties found in many spherical meshes, such as the icosahedral mesh used by GraphCast. However, the HEALPix mesh is particularly amenable for use with SWIN transformers owing to the following properties:
\begin{itemize}
    \item All pixels subdivide into 4 pixels, with each subpixel correspoding to the North, South, East, and West quadrants of the parent pixel.
    \item Pixels have well-defined neighbors along 8 cardinal directions: North-East, North-West, South-East, South-West, North, South, East, and West.
\end{itemize}
Given a refinement level $n$, and window size parameter $w$, shifted windows can be constructed through the following steps:
\begin{enumerate}
    \item Split each coarse pixel at the refinement level $n-w$ into sub-pixels at the refinement level $n-w+1$. As the pixels at level $n-w$ define the original windows, these sub-pixels represent the North, South, East, and West quadrants of the original windows.
    \item Merge each South sub-pixel with its South-West, South, and South-East neighbors to create a new window. All 3 neighbors are guaranteed to be from different windows, as they are further south of the original South sub-pixel.
    \item Similarly, merge each North sub-pixel with its North-West, North, and North-East neighbor, and likewise for the East and West quadrant sub-pixels.
    \item Once this split-merge scheme is applied to the entire mesh at refinement level $n-w$, all windows have been shifted by exactly one quadrant south and west, ensuring that each new window consists of quadrants from 4 separate original windows.
\end{enumerate}
The resulting shifted windows can be visualized as in figure \ref{fig:hpnwindow}. As in a rectilinear mesh, each shifted window on the HEALPix mesh comprises of four quadrants from four neighboring original windows. This allows cross-window connections to be learned in the HEALPix mesh in the same way as in the rectilinear mesh. The polar regions, marked in blue in figure \ref{fig:hpnwindow}, remain undistorted as the patches and windows near the poles are of the same size as those near the equator.

Because all windows contain the same number of mesh nodes, and all mesh nodes represent equal areas in a HEALPix mesh, all pixels perform local self-attention in windows of equal "true" areas on the sphere. In contrast, on a rectilinear grid, even though all windows contain the same number of mesh nodes, windows near the poles represent smaller areas which reduces the "true" context for self-attention near the poles. We note that there are 8 locations where a central HEALPix face meets with 2 northern or 2 southern HEALPix faces, and at these meeting points the shifted windows have only 3 quadrants each, marked in red in figure \ref{fig:hpnwindow}.
\subsection{Efficiencies from the spherical mesh}
\label{section:efficiency}
We can quantify the benefits of using the spherical mesh for ViT models by comparing the number of nodes in similarly refined HEALPix and rectilinear meshes. Pangu-Weather and FuXi, which use SWIN transformers on a rectilinear mesh, use a patch of size 4x4 to reduce the original image from 721x1440 pixels to a rectilinear mesh of size 180x360 (or 64,800 nodes). As the original resolution is at 0.25\textdegree, each node on this mesh represents an area of 1\textdegree x1\textdegree at the equator, and much smaller areas near the poles. Pangu-Weather uses a window of shape $6 \times 12$ to perform local self-attention, thus constructing 900 windows of 72 nodes each.

On a HEALPix mesh, using a refinement level of $n=2^6$, we obtain a mesh with 49,152 nodes, with each node representing an area of 0.91\textdegree{} $\times$ 0.91\textdegree{} everywhere on the sphere. A window parameter $w=3$ results in each window having $(2^3)^2=64$ nodes, with 768 windows covering the entire mesh.

Compared to the rectilinear mesh, the HEALPix mesh has 25\% fewer nodes while also having each node represent a smaller area. 15\% fewer windows are needed to cover the mesh, while being of a similar size to the windows in the rectilinear mesh. As all HEALPix windows have equal area, all mesh nodes attend on equal amounts of input context. In other words, the HEALPix mesh needs fewer nodes to represent data at a higher resolution, which in turn leads to fewer windows to perform attention over, thus lowering memory and compute footprint.
\section{HEAL-ViT}
Having described the HEALPix mesh and how windows are constructed and shifted, we now describe overall model architecture. The HEAL-ViT model follows the encoder-processor-decoder structure as other MLWPs:
\begin{enumerate}
  \item The encoder uses a simple graph network to map the longitude-latitude grid to a HEALPix mesh.
  \item The processor uses SWIN transformers to learn interactions between mesh nodes.
  \item The decoder finally maps the processed HEALPix mesh back to a longitude-latitude grid using a simple graph network.
\end{enumerate}
\subsection{Encoder}
\begin{figure}
  \centering
  \includegraphics[width=0.5\linewidth]{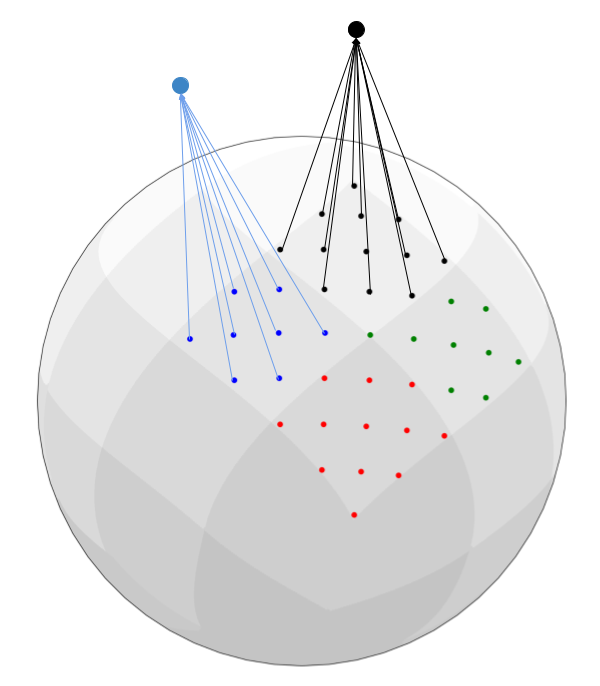}
  \caption{\textbf{Mapping a longitude-latitude grid to HEALPix mesh} Each mesh node is connected to all grid nodes lying within the quadrilateral the mesh node represents. Here, mesh nodes are represented by the floating blue and black dots, while grid nodes are the points located on the sphere.}
  \label{fig:encoder}
\end{figure}

Similar to GraphCast, we use a bipartite graph $\mathcal{G}(\mathcal{V}^{G}, \mathcal{V}^{M}, \mathcal{E}^{G2M})$ to embed the longitude-latitude grid on to a HEALPix mesh. $\mathcal{V}^{G}$ is the set of all grid nodes on the 721x1440 sized longitude-latitude grid. $\mathcal{V}^{M}$ is the set of all mesh nodes on a HEALPix mesh at refinement level $n=7$ with $196,608$ nodes. As each HEALPix mesh node represents a curvilinear quadrilateral area on the sphere, each mesh node is connected to all grid nodes lying within that quadrilateral, as shown in figure \ref{fig:encoder}. $\mathcal{E}^{G2M}$ is the set of directed edges connecting grid nodes to mesh nodes, represented by 32-dimensional embeddings. As each grid node is connected to exactly one mesh nodes, there is one edge per grid node, leading to 1,038,240 edges.\par
Each grid node $\textbf{v}_{i}^{G}$ is represented by the features $\textbf{v}^{G,\text{features}}_{i} = \textbf{[}\textbf{x}_{i}^{t-1}, \textbf{x}_{i}^{t}, \textbf{c}_{i}\textbf{]}$, where $\textbf{x}_{i}^{t}$ is the time-dependent weather state  and $\textbf{c}_{i}$ is the set of constant features for the grid node. Each grid node, and its corresponding edge, is first embedded into a latent space of dimension $C$ using a single feed-forward layer followed by GeLU\cite{hendrycks2016gaussian} activation function:
\begin{equation}
\begin{split}
  \textbf{v}_{i}^{G'} &= \textbf{[}
        e^{\text{G2M}}_{v^{G}_{s} \rightarrow v^{M}_{i}}, \textbf{v}^{G,features}_{i}\textbf{]} \\
  \textbf{v}_{i}^{G} &= \text{GeLU}(\text{W}^{\text{embedder}}\textbf{v}^{G'}_{i} + \text{b}^{\text{embedder}})
\end{split}
\end{equation}
Mesh nodes are computed by aggregating the informating from incoming edges, followed by layer normalization\cite{ba2016layer}. The aggregated information is embedded into a latent space of dimension $C$ using a single feed-forward layer:
\begin{equation}
\begin{split}
  \textbf{v}_{i}^{M'} &= \text{LayerNorm}(\sum_{e^{\text{G2M}}_{v^{G}_{s} \rightarrow v^{M}_{i}}}\textbf{v}^{G}_{s})\\
  \textbf{v}_{i}^{M} &= \text{W}^{\text{G2M}}\textbf{v}_{i}^{M'} + \text{b}^{\text{G2M}}
\end{split}
\end{equation}
In summary, the encoder consumes an image of size $N_{\text{features}} \times 721 \times 1440$ and produces a mesh of size $196608 \times C$.

\subsection{Processor}
\begin{figure}
    \centering
    \includegraphics[width=0.9\linewidth]{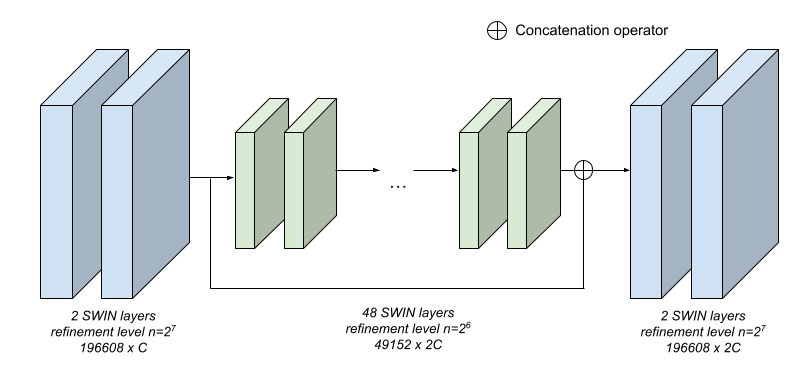}
    \caption{The processor stack for the HEAL-ViT model.}
    \label{fig:processor}
\end{figure}

Once the grid has been embedded into a mesh of size $196608 \times C$, the processor uses a U-Net architecture to learn interactions between the mesh nodes. The encoded mesh is first processed by 2 SWIN blocks, and then coarsened using a down-sample block to a mesh with refinement level $n=2^{6}$ with 49,152 nodes. The coarse mesh of size $49152 \times 2C$ is processed by 48 SWIN blocks and then upsampled back to the original mesh of size $196608 \times C$. A skip connection is used to concatenate the input of the down-sample block to the output of the up-sample block, which is then processed by 2 SWIN blocks. As with typical SWIN transformers, every alternate SWIN block shifts the windows within which local self-attention is performed. The entire processor stack is illustrated in figure \ref{fig:processor}.\par
\textbf{SWIN blocks} We use ViT blocks as described in the ViT-22B\cite{dehghani2023scaling} architecture, which transforms an input $x$ into output $y$ through the following steps:
\begin{equation}
\begin{split}
  y' &= \text{LayerNorm}(x) \\
  y &= x + \text{MLP}(y') + \text{Attention}(y')
\end{split}
\end{equation}
Following the ViT-22B architecture, we do not use biases in the QKV projections, and perform layer normalization on the query and key vectors before computing attention weights:
\begin{equation}
  \text{AttentionWeights} = \text{softmax}\big[\frac{1}{\sqrt{d}}
      \text{LN}(XW^Q)(\text{LN}(XW^K))^T
  \big]
\end{equation}

\textbf{Downsample block} The downsample block coarsens the mesh from $196608 \times C$ to a mesh of size $49152 \times 2C$, effectively making the mesh one refinement level coarser. The downsample block uses a simple graph network on a bipartite $\mathcal{G}(\mathcal{V}^{M_f}, \mathcal{V}^{M_c}, \mathcal{E}^{M_f2M_c})$, where $\mathcal{V}^{M_f}$ is the set of all nodes in the fine mesh (refinement level $2^7$), $\mathcal{V}^{M_c}$ is the set of all nodes in the coarse mesh (refinement level $2^6$), and $\mathcal{E}^{M_f2M_c}$ connects the fine mesh nodes to coarse mesh nodes. Because HEALPix meshes are refined hierarchically, each fine mesh node is connected to exactly one "parent" coarse mesh node. Connecting edges are represented by learnable 32-dimensional embeddings. Each coarse mesh node is updated by first aggregating incoming information from its child nodes, followed by layer normalization. A feed-forward layer then embeds the aggregated information into a latent space of dimension $2C$:
\begin{equation}
\begin{split}
  \textbf{v}_{i}^{M_c'} &= 
    \text{LayerNorm}(\sum_{e^{\text{M}_{f}\text{2M}_{c}}_{v^{M_f}_{s} \rightarrow v^{M_c}_{i}}}\textbf{[}
    e^{\text{M}_{f}\text{2M}_{c}}_{v^{M_f}_{s} \rightarrow v^{M_c}_{i}}, \textbf{v}^{M_f}_{s}\textbf{]}) \\
    \textbf{v}_{i}^{M_c} &= \text{W}^{\text{M}_{f}\text{2M}_{c}}\textbf{v}_{i}^{M_c'} + \text{b}^{\text{M}_{f}\text{2M}_{c}}
\end{split}
\end{equation}
As each coarse node contains 4 fine nodes, the downsample block coarsens 4 fine nodes of dimension $C$ into one coarse node of dimension $2C$, similar to the downsample block used in Pangu-Weather and FuXi.\par
\textbf{Upsample block} The upsample block inverts the procedure of the downsample block, and refines the mesh from size $49152 \times 2C$ to $196608 \times C$. Like the downsample block, upsample block uses a simple graph network on a bipartite graph $\mathcal{G}(\mathcal{V}^{M_c}, \mathcal{V}^{M_f}, \mathcal{E}^{M_c2M_f})$, where $\mathcal{V}^{M_f}$ is the set of all nodes in the fine mesh, $\mathcal{V}^{M_c}$ is the set of all nodes in the coarse mesh, and $\mathcal{E}^{M_c2M_f}$ connects the coarse mesh nodes to fine mesh nodes. We connect each node in the fine mesh to the four nearest nodes in the coarse mesh, and edges are represented using a learnable 32-dimensional embedding. Each fine mesh node is updated using a single feed-forward layer that aggregates the incoming information from its four parent nodes $v_{s}^{M_c}$:
\begin{equation}
\begin{split}
  \textbf{v}_{i}^{M_f'} &= \text{LayerNorm}(\sum_{e^{\text{M}_{f}\text{2M}_{f}}_{v^{M_c}_{s} \rightarrow v^{M_f}_{i}}}\textbf{[}
    e^{\text{M}_{c}\text{2M}_{f}}_{v^{M_c}_{s} \rightarrow v^{M_f}_{i}}, \textbf{v}^{M_c}_{s}\textbf{]})\\
    \textbf{v}_{i}^{M_f} &= \text{W}^{\text{M}_{c}\text{2M}_{f}}\textbf{v}_{i}^{M_f'} + \text{b}^{\text{M}_{c}\text{2M}_{f}}
\end{split}
\end{equation}
\subsection{Decoder}
Similar to the upsample block, we use a bipartite graph $\mathcal{G}(\mathcal{V}^{G}, \mathcal{V}^{M}, \mathcal{E}^{M2G})$ to map the HEALPix mesh of size $196608 \times 2C$ back to the longitude-latitude grid. $\mathcal{V}^{G}$ is the set of all grid nodes on the 721x1440 sized longitude-latitude grid, $\mathcal{V}^{M}$ is the set of all mesh nodes, and $\mathcal{E}^{M2G}$ is the set of directed edges connecting mesh nodes to grid nodes. As with the upsample block, each grid node is connected to the four nearest HEALPix mesh nodes, with edges being represented by a 32-dimensional learnable embeddings. Each grid node is updated by aggregating information from the four nearest mesh nodes, followed by a layer normalization. A feed-forward layer then transforms the aggregated information into a 54-dimensional output per grid node, matching the number of predicted variables.\par
\begin{equation}
\begin{split}
  \textbf{v}_{i}^{G'} &= \text{LayerNorm}(\sum_{e^{\text{M2G}}_{v^{M}_{s} \rightarrow v^{G}_{i}}}\textbf{[}
    e^{\text{M2G}}_{v^{M}_{s} \rightarrow v^{G}_{i}}, \textbf{v}^{M}_{s}\textbf{]})\\
    \textbf{v}_{i}^{G} &= \text{W}^{\text{M2G}}\textbf{v}_{i}^{G'} + \text{b}^{\text{M2G}}
\end{split}
\end{equation}

\subsection{Normalization and network parameterization}
\subsection{Input and output normalization} Before being fed to the encoder, all input variables in table \ref{tab:weatherfeats} are normalized to zero mean and unit variance, using the mean and variance statistics computed over the entire training period. The target is similarly normalized, and the network directly predicts the normalized target as the output. Mean and variance statistics are computed separately for each variable and level.
\subsection{Neural network parametrization} We choose a value of $C=256$ for the output of the encoder, i.e., the encoder embeds the input grid from $110 \times 721 \times 1440$ to a mesh of size $196608 \times 256$. The downsample block in the processor thus coarsens the mesh from $1966608 \times 256$ to a mesh of size $49152 \times 512$. The MLP within each SWIN layer has one hidden layer of size 4*D (where $D$=$C$ or $2C$ depending on which mesh the SWIN layer acts on). The self-attention module of the SWIN block uses a head dimension $d=32$, leading to 8 or 16 heads depending on the mesh processed by the SWIN block. All edges are represented by learnable 32-dimensional embeddings.

\section{Training details}
\subsection{Training objective}
We optimize HEAL-ViT to minimize the latitude-weighted L1 loss:
\begin{equation}
  \text{L1} = \frac{1}{C \times H \times W}
  \sum_{c=1}^{C}
  \sum_{h=1}^{H}
  \sum_{w=1}^{W}
  w(h) | \phi(X^{t}, X^{t-1})_{c,h,w} - X^{t+1}_{c,h,w}|
\end{equation}
where $C$ is the number of predicted variables, and $H$ and $W$ are the number of latitudes and longitudes in the output grid. For auto-regressive training (described below), the L1 loss from all time-steps is averaged to obtain an overall training loss. $w(i)$ represents the weight of the grid cell at the $i$-th latitude, as detailed in Section \ref{section:latweight}.
\subsection{Training and testing split}
We train on ERA5 data from 1992-2018, using data from 2019 as the development set and data from 2020 as the test set. The development set is used for fine-tuning hyper-parameters, and the test set is used only for the final evaluation of the model.
\subsection{Curriculum training schedule}
We first pre-train the HEAL-ViT model to predict a single step $X^{t+1} = \phi(X^t, X^{t-1})$ for 80,000 steps. Once the model has been trained to predict the immediate next step, we follow GraphCast and use an auto-regressive training regime to improve the model performance beyond the first step as described in section \ref{section:problem}. Compared to GraphCast, which is trained for up to 12 auto-regressive steps, and FuXi, which is trained for 20 auto-regressive steps, HEAL-ViT is fine-tuned only for 4 auto-regressive steps with the following training curriculum:
\begin{itemize}
  \item 2 autoregressive steps, trained for 4000 training steps
  \item 3 autoregressive steps, trained for 4000 training steps
  \item 4 autoregressive steps, trained for 4000 training steps
\end{itemize}
\par
\subsection{Hyperparameters}
We use the AdamW optimizer \cite{kinga2015method,loshchilov2017decoupled} (beta1 = 0.9, beta2 = 0.95) with a weight decay of 0.1. For the pre-training phase, the learning rate is initialized to 2.5e-4, and decayed using a cosine schedule. The auto-regressive fine-tuning is performed with a fixed learning rate of 1e-7. We use 8 A100 GPUs, placing 1 example per GPU, and use activation checkpointing\cite{chen2016training} to reduce the memory footprint needed to train the models. We use PyTorch \cite{paszke2019pytorch} and xarray\cite{hoyer2016xarray} to process the data and train the models. The pre-training phase takes roughly 2.5 days, while the fine-tuning process takes an additional 2 days.

\section{Evaluation results}
We evaluate HEAL-ViT's forecast performance on 2020, and compare it to both ECMWF's IFS and other MLWPs, using the WeatherBench2\cite{rasp2023weatherbench} benchmarking suite to perform the evaluations. Like other MLWPs, HEAL-ViT is initialized to start at 00UTC each day in 2020 using ERA5 data.
\subsection{Baselines}
\subsubsection{NWP Baselines}
ERA5 uses a longer assimilation window at 00UTC, which makes it a higher quality state to initialize a weather model with. As all MLWPs, including HEAL-ViT, are initialized using the ERA5 data at 00UTC on each day, HEAL-ViT has an advantage over the operational IFS (which does not have access to ERA5 data), making a like-for-like comparison between MLWPs and IFS difficult. WeatherBench2 provides "ERA5 forecasts", a set of IFS forecasts initialized using ERA5 data, which provide a like-for-like baseline to compare MLWPs against, as all models now start from the same start state and can be compared against ERA5 as ground truth. We thus use IFS forecasts initialized using ERA5 data as a NWP baseline, labeling them ERA5-IFS in the comparisons below.\par
\begin{figure}
  \captionsetup{singlelinecheck = false, format= hang, justification=raggedright}
  \centering
  \includegraphics[width=\linewidth]{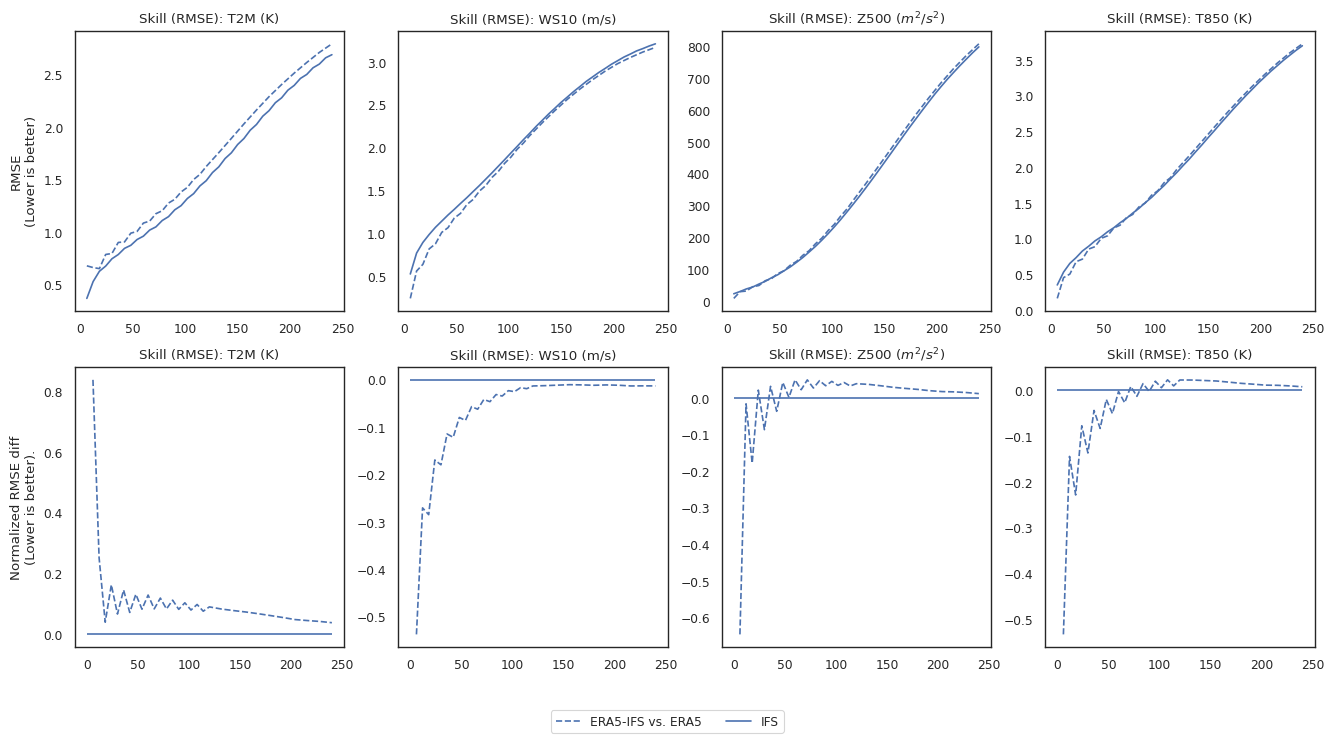}
  \caption{\textbf{RMSE comparison, IFS vs. ERA5-IFS.} The operational IFS forecast (solid blue) has lower RMSE than the IFS model initialized with ERA5 data (dashed blue) for T2M, while the reverse is true for WS10. ERA5-IFS performs better for shorter lead times for Z500 and T850. Normalized RMSE difference uses operational IFS model as the baseline.}
  \label{fig:era5ifsifs}
\end{figure}

Figure \ref{fig:era5ifsifs} shows the performance difference of ERA5-IFS against the operational IFS, as well as the normalized difference of RMSE, calculated as the relative change compared to a baseline model $B$: $(\text{RMSE}_{A} - \text{RMSE}_{B})/\text{RMSE}_{B}$. We note that the operational IFS consistently performs better for T2M forecasts, while performing worse for WS10 forecasts over the entire forecast horizon. For Z500 and T850, ERA5-IFS performs better for shorter lead times compared to IFS. For all variables, we note that any performance difference reduces, but is still present over longer lead times.
\subsubsection{MLWP baselines}
We use GraphCast, Pangu-Weather, and FuXi as MLWP baselines, sourcing the evaluation statistics from WeatherBench2. While all MLWPs are initialized using ERA5 data and evaluated over 2020, GraphCast is initialized with start states at 00UTC and 12UTC, while the other MLWPs are initialized at 00UTC.
\subsection{Metrics}
\label{section:latweight}
All metrics are computed using an area-weighted average over the entire grid to account for the fact that grid cells represent smaller areas towards the poles compared to at the equator. Weights are computed for each latitude as below:
\begin{equation}
  w(i) = \frac{
    \sin(\theta_{i}^{u}) - \sin(\theta_{i}^{l})
  }{
    \frac{1}{I}\sum_{i}^{I}(\sin(\theta_{i}^{u}) - \sin(\theta_{i}^{l}))
  }
\end{equation}
where $\theta_{i}^{u}$ and $\theta_{i}^{l}$ are the upper and lower latitude bounds for the grid cell with latitude index $i$. In the defintions below, $f_{t,l,i,j}$ indicates the forecasted value for a variable at time $t$, level $l$, and grid cell $(i,j)$. The observed value for the same time, level, and grid cell is represented by $o_{t,l,i,j}$.\par
\subsubsection{Root Mean Squared Error (RMSE)} Consistent with WMO and ECMWF, WeatherBench2 defines the RMSE for each variable and level as
\begin{equation}
  \text{RMSE}_{l} = \sqrt{
    \frac{1}{TIJ}
    \sum_{t}^{T}\sum_{i}^{I}\sum_{j}^{J}
    w(i)(f_{t,l,i,j} - o_{t,l,i,j})^2
  }
\end{equation}
\subsubsection{Anomaly Correlation Coefficient (ACC)} The ACC represents the correlation between forecasted and observed anomalies with respect to climatology $c_{t,l,i,j}$:
\begin{equation}
\begin{split}
  f'_{t,l,i,j} &= f_{t,l,i,j} - c_{t,l,i,j} \\
  o'_{t,l,i,j} &= o_{t,l,i,j} - c_{t,l,i,j} \\
  \text{ACC}_{l} &= \frac{1}{T}\sum_{t}^{T}
    \frac{
      \sum_{i}^{I}\sum_{j}^{J} w(i) f'_{t,l,i,j} o'_{t,l,i,j}
    }{
      \sqrt{
        \sum_{i}^{I}\sum_{j}^{J}w(i) {f_{t,l,i,j}'}^2
        \sum_{i}^{I}\sum_{j}^{J}w(i) {o_{t,l,i,j}'}^2
      }
    }
\end{split}
\end{equation}
Climatology represents the average weather for a specific day of year and time of day, calculated over long historical period. We refer to WeatherBench2\cite{rasp2023weatherbench} for details on how climatology is computed.
\subsubsection{Bias} Bias is computed simply as
\begin{equation}
  \text{Bias}_{l} =\frac{1}{TIJ}
    \sum_{t}^{T}\sum_{i}^{I}\sum_{j}^{J}
    w(i)(f_{t,l,i,j} - o_{t,l,i,j})
\end{equation}
MLWPs have shown a tendency to accumulate negative biases strongly, which can impact the ability to forecast intensity of tropical cyclones over longer lead times\cite{demaria2024cyclone}, thus making it important to consider bias along with other quality metrics.
\subsubsection{Energy Spectra}
Energy spectra indicate how much variation there is on different spatial scales for each variable and level. More power on smaller scales indicate that forecast fields have better smaller-scale structure, and lesser power indicates blurrier forecasts. In WeatherBench2, zonal spectral energy is calculated separately for lines of constant latitude, and a final spectrum is computed by averaging the zonal spectra between $30\degree < |lat| < 60\degree$. We follow the same procedure for HEAL-ViT.

\section{Results}
We report the results for 2m temperature (T2M), 10m wind speeds (WS10), geopotential at height 500hPa (Z500), and temperature at 850hPa (T850). These variables are some of the most commonly evaluated variables in medium-range forecasting.
\subsection{RMSE}
\begin{figure}
  \captionsetup{singlelinecheck = false, format= hang, justification=raggedright}
  \centering
  \includegraphics[width=\linewidth]{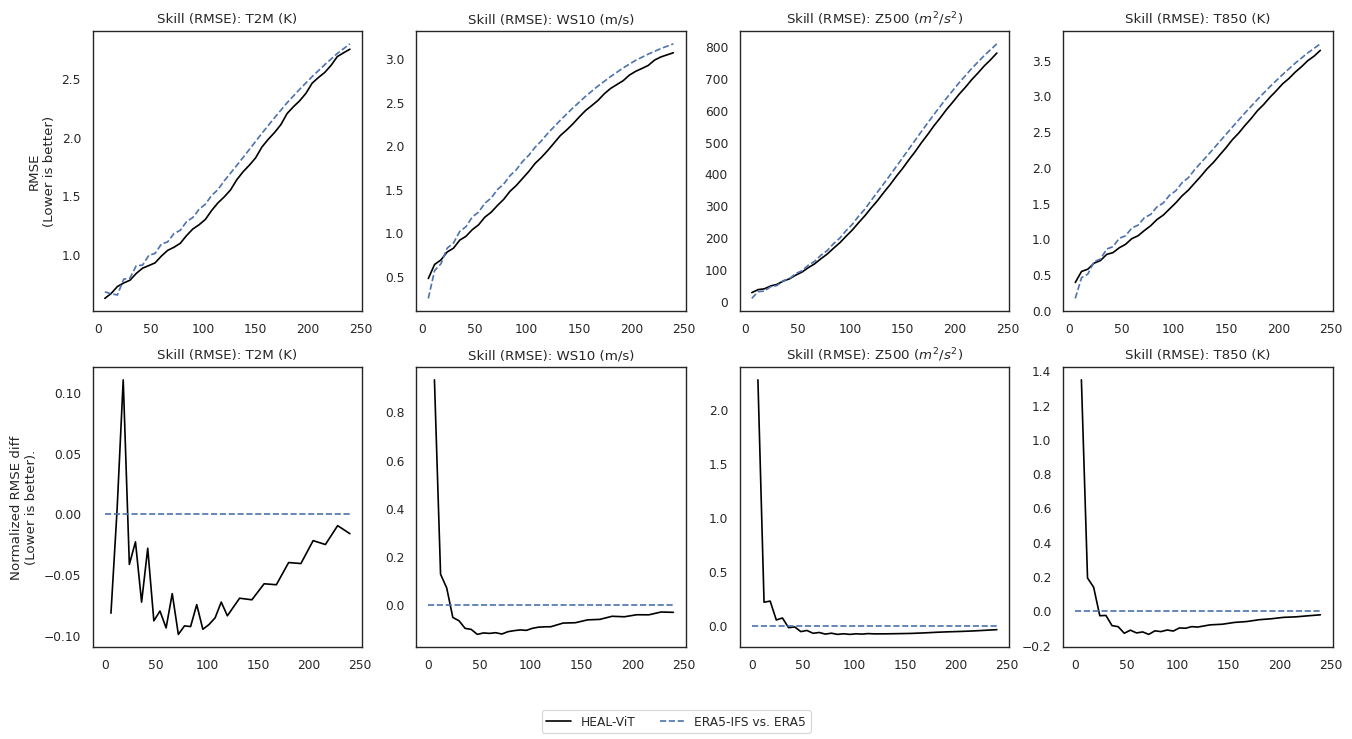}
  \caption{\textbf{RMSE comparison, HEAL-ViT vs. ERA5-IFS.} HEAL-ViT (solid black) has higher RMSE for very short lead times compared to ERA5-IFS (dashed blue), but consistently performs better after 3 to 4 forecast steps.}
  \label{fig:rmse_nwp}
\end{figure}

Figure \ref{fig:rmse_nwp} shows the RMSE of HEAL-ViT and ERA5-IFS for T2M, WS10, Z500, and T850, as well as the normalized difference of RMSE. We observe that for all variables ERA5-IFS has lower RMSE for the first 18 hours (3 forecast steps), after which the HEAL-ViT model consistently shows lower error. We recall that ERA5-IFS strongly outperforms operational IFS for WS10, Z500, and T850 at short lead times, indicating that the IFS model strongly  benefits from having ERA5 data as a start state.\par
\begin{figure}
  \captionsetup{singlelinecheck = false, format= hang, justification=raggedright}
  \centering
  \includegraphics[width=\linewidth]{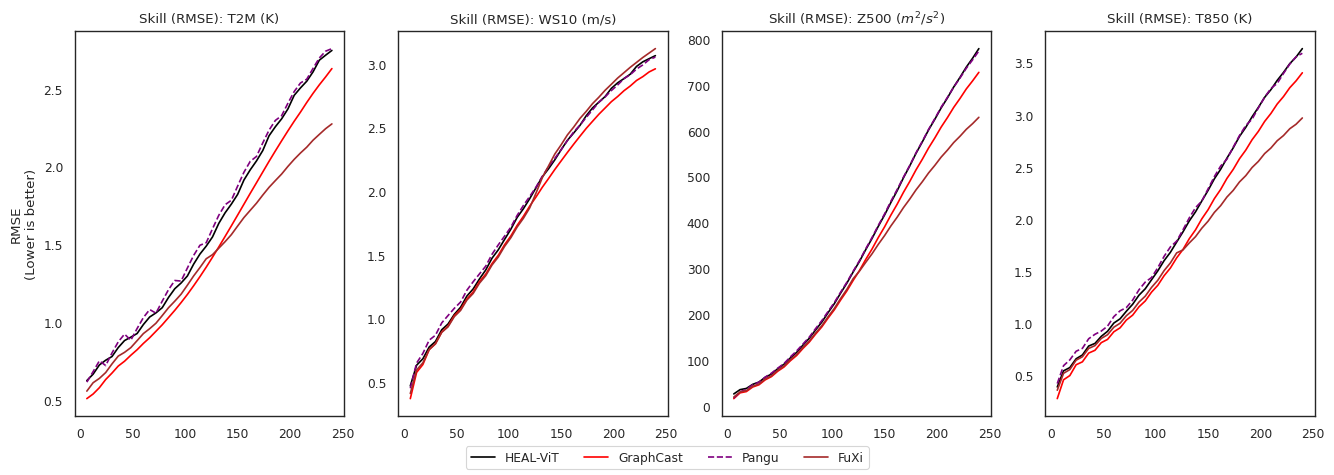}
  \caption{\textbf{RMSE comparison for MLWPs.} HEAL-ViT shows lower RMSE than Pangu-Weather, but higher than GraphCast and FuXi. The exception is WS10 forecasts, where FuXi performs worse after day 5.}
  \label{fig:rmse_mlwp}
\end{figure}

Figure \ref{fig:rmse_mlwp} shows the RMSE of HEAL-ViT and other MLWPs, namely, GraphCast, Pangu-Weather, and FuXi. While HEAL-ViT outperforms Pangu-Weather, we observe that GraphCast and FuXi have lower RMSE than HEAL-ViT (except for WS10, where FuXi has the highest error beyond day 5).

\subsection{ACC}
\begin{figure}
  \captionsetup{singlelinecheck = false, format= hang, justification=raggedright}
  \centering
  \includegraphics[width=\linewidth]{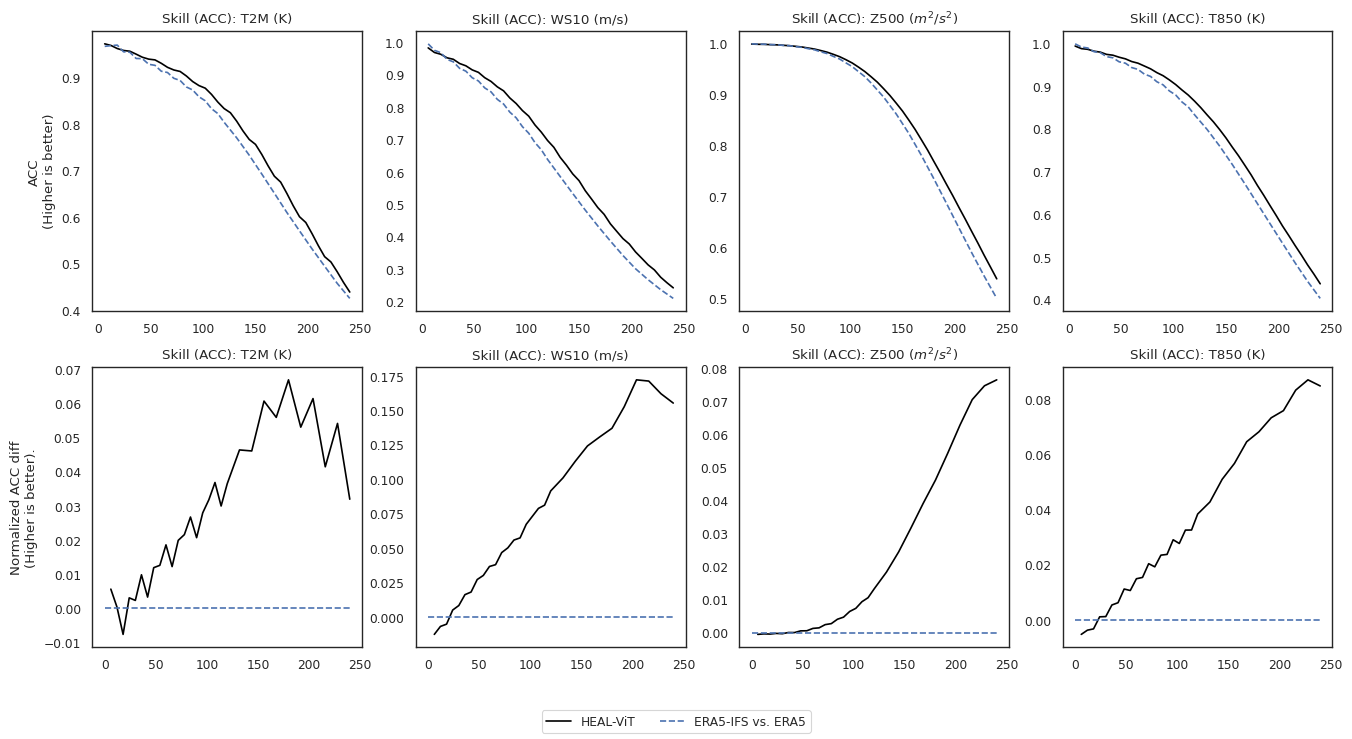}
  \caption{\textbf{ACC comparison, HEAL-ViT vs. ERA5-IFS.} As with RMSE, ERA5-IFS performs well for the very short lead times (3 to 4 steps), after which HEAL-ViT has better performance.}
  \label{fig:acc_nwp}
\end{figure}

Figure \ref{fig:acc_nwp} shows the ACC of HEAL-ViT and ERA5-IFS for the examined variables, as well as the normalized difference of ACC. We observe similar trends for ACC as for RMSE, namely that ERA5-IFS has higher ACC for the first 3 to 4 steps, after which the HEAL-ViT model consistently shows higher ACC.\par
\begin{figure}
  \captionsetup{singlelinecheck = false, format= hang, justification=raggedright}
  \centering
  \includegraphics[width=\linewidth]{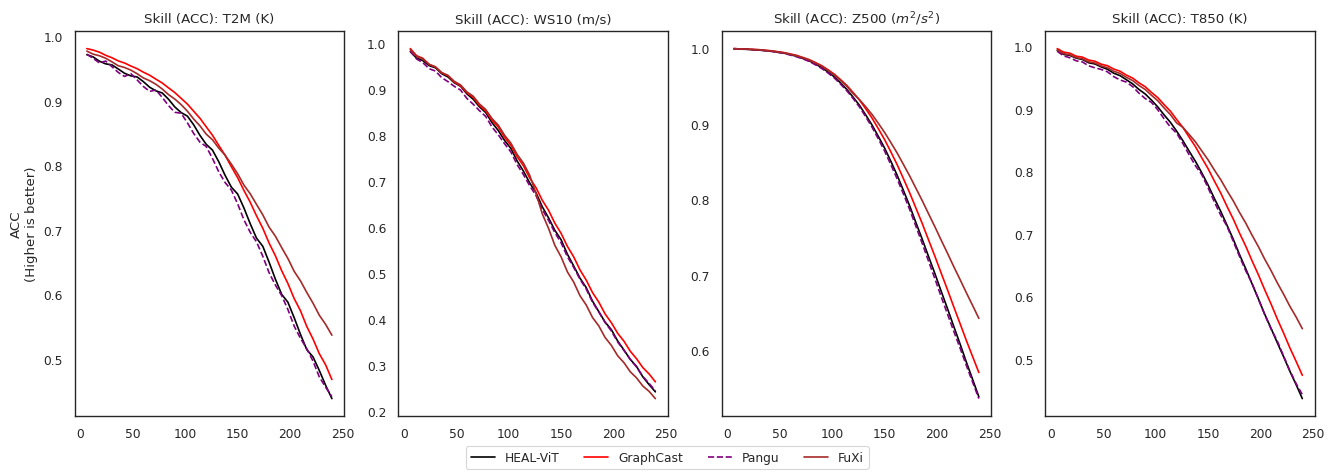}
  \caption{\textbf{ACC comparison for MLWPs.} As with RMSE, HEAL-ViT has better ACC than Pangu-Weather but worse than GraphCast and FuXi. The exception is WS10 forecasts, where FuXi performs worse after day 5.}
  \label{fig:acc_mlwp}
\end{figure}

Figure \ref{fig:acc_mlwp} shows the ACC of HEAL-ViT and other MLWPs, once again showing similar trends as for RMSE. HEAL-ViT outperforms Pangu-Weather, while GraphCast and FuXi have better performance (except for WS10, where FuXi has worse ACC after day 5).

\subsection{Bias}
\begin{figure}
  \captionsetup{singlelinecheck = false, format= hang, justification=raggedright}
  \centering
  \includegraphics[width=\linewidth]{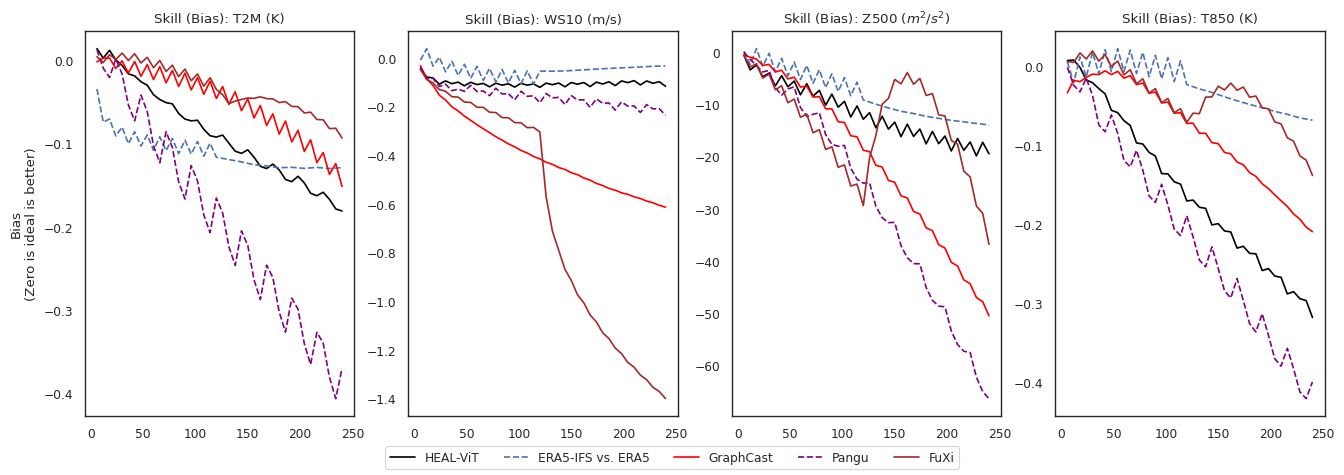}
  \caption{\textbf{Bias comparison for NWP and MLWPs.}}
  \label{fig:bias_mlwp}
\end{figure}

Figure \ref{fig:bias_mlwp} shows the bias for ERA5-IFS and all MLWPs. As bias measures how much a forecast under- or over-estimates a variable, an ideal forecast has zero bias. While all models accumulate negative biases, we note that ERA5-IFS accumulates bias slowly over the duration of the forecast.\par
In contrast, while MLWPs show lower RMSE than ERA5-IFS, they tend to accumulate negative biases more strongly. For instance, Pangu-Weather shows a strong negative bias for T2M, while GraphCast and FuXi show strong negative bias for WS10, Z500, and T850. In particular, FuXi shows discontinuities in bias trends, arising from a separate model being used after day 5 for FuXi forecasts. HEAL-ViT tends to perform better than other MLWPs for bias accumulation, especially evident for T2M, WS10, and Z500, where HEAL-ViT accumulates bias in a similar range as ERA5-IFS.\par
\subsection{Energy spectra}
\begin{figure}
  \captionsetup{singlelinecheck = false, format= hang, justification=raggedright}
  \centering
  \includegraphics[width=\linewidth]{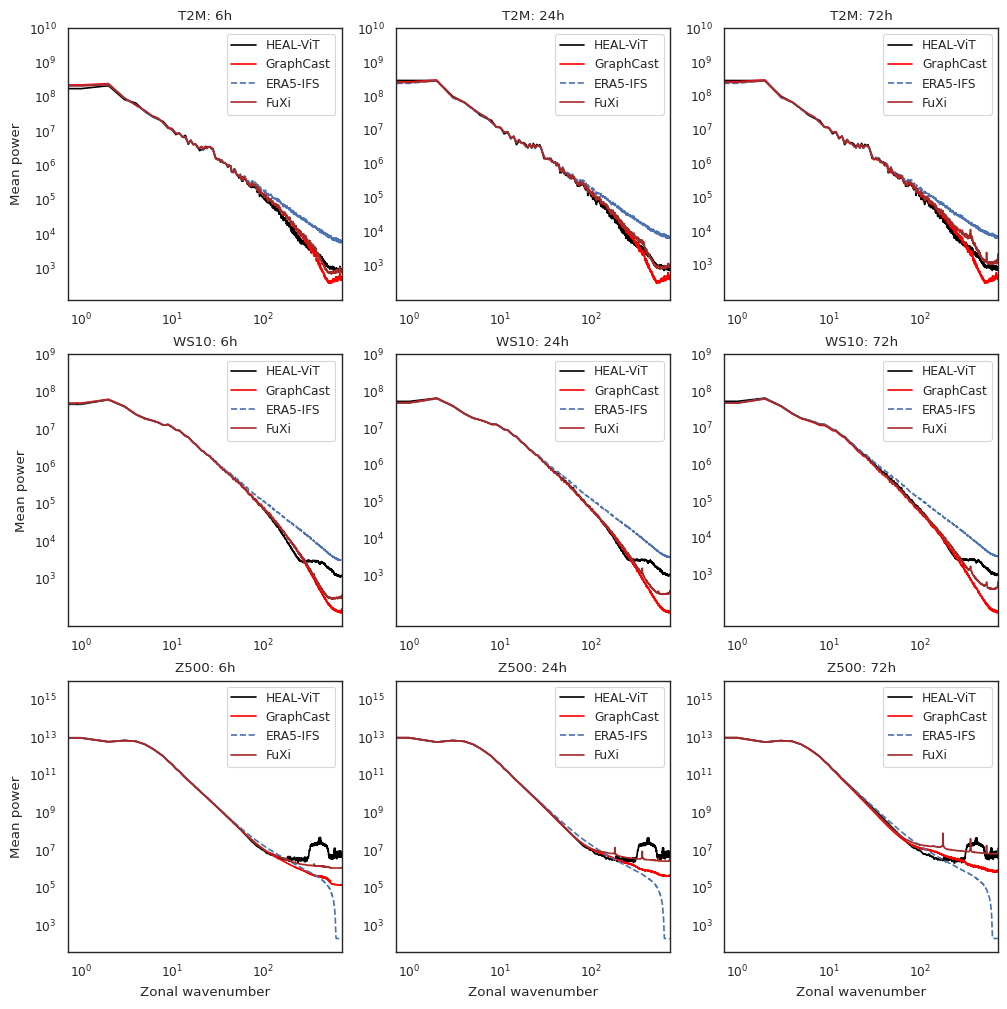}
  \caption{\textbf{Power spectra for T2M, WS10, Z500.}. The power spectra show that ERA5-IFS has better structure at finer scales for T2M and WS10 at all lead times. HEAL-ViT is performs better than other MLWPs in having higher energy at smaller scales for WS10 and Z500, and comparable energy for T2M.}
  \label{fig:spectra}
\end{figure}

Zonal energy spectra indicate how blurry a forecast is, with more energy on smaller scales indicating that a forecast has better small-scale structure. MLWPs tend to have blurrier forecasts than NWPs, arising from the choice of MAE or MSE as a loss function. We show the zonal energy spectra for lead times of 6 hours, 1 day, and 3 days, after which most MLWPs have relatively constant spectra. The spectra for ERA5-IFS are also shown, which are nearly constant across all lead times. Figure \ref{fig:spectra} shows the energy spectra for the MLWPs and ERA5-IFS. We see that for T2M, all MLWPs have similar spectra (though GraphCast tends to be blurrier at all lead times), and ERA5-IFS has much higher power at smaller scales. For WS10, HEAL-ViT has much higher energy than FuXi and GraphCast, though still lower than ERA5-IFS. For Z500, MLWPs have higher energy on smaller scales than ERA5-IFS, and the difference increases with longer lead times. Here as well, HEAL-ViT has more energy on smaller scales than FuXi and GraphCast, though by day 3 FuXi and HEAL-ViT are comparable. We also note that the spectra for HEAL-ViT are relatively constant across all lead times, similar to ERA5-IFS.\par
The spectra also reveal that the choice of encoder has an impact on the forecasted fields. FuXi, which uses a regular 2D patch in the encoder, has spikes in the spectra possibly as a result of patch boundaries. The spikes are also accentuated with longer lead times, indicating that the impact of any patching mechanism is accentuated when forecasts are produced auto-regressively. HEAL-ViT also exhibits a "bump" at the smaller scale, likely arising from its own encoder and mesh-to-grid architecture.

\section{Conclusions and future work}
By combining the use of graph networks to construct a spherical mesh, and vision transformers to learn interactions between mesh nodes, HEAL-ViT provides an architecture that benefits from unique advantages of GraphCast (a spatially homogeneous mesh) and transformer-based methods (efficient learning of teleconnections). The approach is not specific to the problem of medium-range weather forecasting, and can be generalized to any problems using geo-spatial data on longitude-latitude grids. The reduction in memory footprint discussed in \ref{section:efficiency} is valuable when MLWPs are used in operational settings, where more than just a single MLWP need to run concurrently to produce hourly forecasts for more variables than are included in the current MLWPs. As seen with other MLWPs, HEAL-ViT demonstrates that compared with like-for-like NWP baselines, ML weather models are able to outpeform NWPs on key metrics of forecast quality. Additionally, HEAL-ViT improves upon existing MLWPs through better bias accumulation and reduced blurriness as seen in energy spectra.\par

The performance of HEAL-ViT can likely be further improved by improving the encoder, processor, and decoder components. We use rudimentary graphs for the encoder and decoder, which do not contain any prior knowledge about the absolute relative positions of mesh and grid nodes (as do the graphs used by GraphCast). It is likely that such inductive biases will improve the quality of the encoder and the decoder, and thus the overall model. The processor consists of standard vision transformer blocks that do not incorporate variations that have demonstrated improvements in other applications, e.g., learnable relative position biases, post-layer normalization, scaled cosine attention, scheduled DropPath\cite{larsson2017ultra} etc. These improvements were incorporated in FuXi to dramatically increase the model size (relative to the SWIN model used in Pangu-Weather). It is likely that any improvements made to ViT architectures will improve performance for HEAL-ViT as well. As the HEAL-ViT mesh places nodes on regular latitude rings, other options for learning teleconnections are trivially possible, e.g., by providing adjacent latitude rings as a context window instead of a window of mesh nodes.\par

The lowered memory and compute requirements of HEAL-ViT, along with the ability to use standard vision transformer architectures, helps us readily adopt any advancements made in the field of computer vision. In addition to the improvements described above, we aim to explore the use of HEAL-ViT for related applications in weather forecasting, e.g., super-resolution, probabilistic forecasting, and learning reusable, high-quality representations of weather states. 

\bibliographystyle{apalike}
\bibliography{refs}

\end{document}